\documentclass{article}

\usepackage[preprint]{neurips_2026}

\usepackage[utf8]{inputenc}
\usepackage[T1]{fontenc}
\usepackage[hidelinks]{hyperref}
\usepackage{url}
\usepackage{booktabs}
\usepackage{amsfonts}
\usepackage{amsmath}
\usepackage{amssymb}
\usepackage{microtype}
\usepackage{xcolor}
\usepackage{graphicx}
\usepackage{enumitem}
\usepackage{float}
\title{Targeted Tests for LLM Reasoning: An Audit-Constrained Protocol}

\author{%
Hongmin Li\textsuperscript{1,2}\\[0.25em]
\normalfont\textsuperscript{1}School of Life Science and Technology, Institute of Science Tokyo\\
\normalfont 2-12-1 Ookayama, Meguro-ku, Tokyo 152-8550, Japan\\
\normalfont\textsuperscript{2}Department of Computational Biology and Medical Sciences\\
\normalfont Graduate School of Frontier Sciences, The University of Tokyo\\
\normalfont 5-1-5 Kashiwanoha, Kashiwa-shi, Chiba 277-8561, Japan\\
\normalfont\texttt{lihongmin@edu.k.u-tokyo.ac.jp}\\[0.25em]
\normalfont\small Researcher, School of Life Science and Technology, Institute of Science Tokyo;\\
\normalfont\small Guest Researcher, Department of Computational Biology and Medical Sciences,\\
\normalfont\small Graduate School of Frontier Sciences, The University of Tokyo.\\
\normalfont\small ORCID: \href{https://orcid.org/0000-0003-0228-0600}{0000-0003-0228-0600}
}

\begin{document}

\maketitle

\begin{abstract}
Fixed reasoning benchmarks evaluate canonical prompts, but semantically valid
changes in presentation can still change model behavior. Studies of prompt
variation can reveal such failures, but without audit they can mix genuine
model errors with invalid perturbations, extraction artifacts, and unmatched
search procedures. We propose an audit-constrained protocol for targeted
reasoning evaluation. Prompt variants are generated from a finite component
grammar, rendered deterministically, evaluated under a fixed query budget, and
counted as model errors only after semantic and extraction audit. Within this protocol
we instantiate Component-Adaptive Prompt Sampling (CAPS), a score-based sampler
over prompt components, and compare it with equal-budget uniform component
sampling under the same task bank, renderer, model interface, decoding
settings, and audit procedure. Across three audited slices, the protocol
identifies confirmed model-error prompt keys while excluding formatting and
extraction artifacts, but matched comparisons do not show that CAPS improves
audited yield or unique prompt-key discovery over uniform sampling. The
contribution is methodological: targeted prompt variation can be studied under
a reconstructable, reviewable, budget-matched protocol, and
proxy-guided policies should be judged by audited yield rather than raw
mismatch counts or selected examples alone.

\end{abstract}

\section{Introduction}
Large language models can solve many benchmark reasoning problems
\citep{brown2020language,srivastava2022beyond,wei2022chain}, yet their behavior
remains sensitive to prompt wording, exemplar choice, and minor surface-form
perturbations \citep{kojima2022large,zhao2021calibrate}. Fixed benchmarks
evaluate one canonical presentation of each task. This makes results easy to
compare, but it can miss failures that appear under semantically valid changes
in presentation. Open-ended prompt search can surface such failures, but often
makes it harder to tell whether the perturbation preserved the task and whether
the observed mismatch is a genuine model error.

We study \emph{targeted reasoning tests} under these constraints. A targeted test is a
compact set of candidate prompt variants generated from a finite
grammar with documented semantic-validity criteria and evaluated under a fixed
query budget. The aim is not to improve a model's answers, but to identify
semantically valid prompt variants under which the same underlying task is
answered incorrectly. This requires checking two things separately: whether the
prompt perturbation preserved the task, and whether the observed answer
mismatch reflects a model error rather than a parsing or formatting artifact.

Our main contribution is an audit-constrained protocol for making targeted
prompt variation inspectable and comparable. Under the protocol, every queried
candidate is tied to a frozen task bank, a deterministic rendering path, and
explicit validity and audit decisions, so that automatic proxy mismatches are
not confused with adjudicated model-error claims. The protocol can support
different construction policies. In this paper, we instantiate
Component-Adaptive Prompt Sampling (CAPS), a score-based categorical sampler
over prompt components, as one candidate policy and compare it with
equal-budget uniform component sampling under the same task bank, renderer,
query budget, and audit procedure. CAPS does not use model gradients or search
arbitrary token strings; it is evaluated here as a diagnostic sampling policy
within the protocol, not as a method for improving prompts.

This paper makes three contributions:
\begin{itemize}[leftmargin=*]
    \item an audit-constrained formulation of targeted reasoning-test
    construction under structural-validity, semantic-validity,
    extraction-validity, and query-budget constraints;
    \item a matched comparison framework with CAPS as a proxy-guided candidate
    policy and uniform component sampling as an equal-budget control under the
    same task bank, renderer, model interface, decoding settings, query budget,
    and audit procedure;
    \item an audited diagnostic study showing that the probe banks contain
    model-facing prompt variants that survive review as confirmed model-error
    evidence, while finding no matched-budget evidence that CAPS improves
    audited yield or unique prompt-key discovery over uniform component
    sampling.
\end{itemize}

The empirical question is therefore not whether CAPS can be illustrated with
selected successes, but whether targeted prompt variation yields reviewable
evidence and whether a proxy-guided policy improves audited discovery over a
matched uniform baseline. The present evidence supports the protocol and the
existence of specific audited variant-level errors, but does not show that
adaptive component scoring is superior to uniform sampling.

\section{Related Work}
Benchmarks for language-model reasoning have grown from broad multitask
collections to increasingly difficult suites that probe compositional,
mathematical, and symbolic behavior \citep{brown2020language,srivastava2022beyond}.
Prompting methods such as chain-of-thought and zero-shot chain-of-thought show
that elicitation format can substantially change measured reasoning performance
\citep{wei2022chain,kojima2022large}. These results motivate evaluations that
record not only model accuracy, but also the prompt conditions under which that
accuracy is observed.

Prompt sensitivity is also a known source of instability. Calibration and
contextual effects can make predictions depend on prompt wording or exemplar
choice even when the underlying task is unchanged \citep{zhao2021calibrate}.
Adversarial and behavioral testing work provides a complementary view.
CheckList-style tests organize expected behaviors into targeted capabilities
\citep{ribeiro2020beyond}; contrast sets ask whether labels remain stable under
small, meaning-preserving edits \citep{gardner2020evaluating}; adversarial
triggers and red-teaming seek inputs that expose brittle or undesirable
behaviors \citep{wallace2019universal,perez2022red}. These lines of work
establish that targeted inputs are useful, but they usually do not define a
single artifact contract that ties each sampled prompt to a frozen task bank,
deterministic rendering path, raw model response, automatic mismatch signal,
matched construction-policy control, and resolved audit label.

This work differs from fixed benchmark construction and manual behavioral
testing by making targeted prompt-test construction an auditable evaluation
object. The goal is not to improve model outputs, but to produce compact prompt
batches that preserve task semantics, expose candidate reasoning failures, and
retain enough evidence for semantic and extraction review. This emphasis also
separates the present protocol from reports that list selected brittle examples:
each candidate is generated from a frozen task bank and deterministic grammar,
queried under a fixed budget, preserved with the raw model response, and counted
as evidence only after audit. Relative to prior behavioral tests, the protocol
fixes six audit axes: frozen task-bank version, deterministic prompt grammar,
matched CAPS-versus-uniform comparison, raw-response preservation,
proxy-versus-audit separation, and unique-key accounting.

The matched-policy comparison is included to test a modest evaluation
hypothesis, not to make the sampler the contribution. A proxy-guided component
sampler is a natural candidate policy for targeted-test construction, but the
paper treats its automatic mismatch signal as candidate routing rather than as a
failure label. The equal-budget uniform component sampler is therefore part of
the evaluation design: it checks whether a more directed policy actually
improves audited yield or unique audited prompt-key discovery under the same
task bank, grammar, model calls, and audit path. The contribution is the
targeted-test protocol and artifact contract rather than a claim about a
particular sampling policy.

\section{Method}
We study targeted reasoning-test construction under an audit-constrained
protocol. The protocol defines a controlled measurement setting: it fixes a
task bank and a finite perturbation grammar, records every queried candidate,
and separates automatic routing signals from audited model-error claims.
Within this protocol, Component-Adaptive Prompt Sampling
(CAPS) is a proxy-guided policy for sampling from a finite prompt-component
grammar. It operates over discrete auditable prompt components rather than
arbitrary token strings, and it is used here for targeted test construction
rather than prompt optimization.

\paragraph{Task bank and candidate space.}
Let $\mathcal{T}$ be a finite bank of reasoning tasks. Each task is a tuple
\[
    t = (id, f, q, a, v),
\]
where $id$ is a stable identifier, $f$ is a task family, $q$ is the canonical
question, $a$ is the canonical answer together with its normalization or
equivalence rule, and $v$ is a set of validity tags. We also fix a finite
prompt-component space $\mathcal{C}$ and a deterministic renderer
$R:\mathcal{T}\times\mathcal{C}\rightarrow\mathcal{P}$. A queried candidate is
\[
    x=(t,c,p), \qquad p=R(t,c),
\]
where $c$ is a component assignment and $p$ is the rendered prompt. Because the
renderer is deterministic, every queried prompt is reconstructable from the
task-bank snapshot and the stored component assignment.

In our experiments,
\[
    \mathcal{C} = \mathcal{C}_{format}
    \times \mathcal{C}_{distractor}
    \times \mathcal{C}_{reading}.
\]
The format group controls whether the model should answer directly, explain
briefly, or check a tempting shortcut. The distractor group adds no distractor,
an irrelevant number, or a plausible shortcut sentence. The reading-cue group
prepends canonical, reversed, or interleaved reading instructions without
permuting the task sentence itself. This grammar is intentionally small so that
sampled prompts remain auditable and manual semantic review remains feasible.

\paragraph{Validity contract.}
Validity is part of the measurement object rather than a post-hoc annotation. We
distinguish structural validity from semantic validity. The automatic predicate
$V_{\mathrm{auto}}(t,c)$ checks requirements such as prompt length, task-anchor
retention, and answer-key presence. The semantic predicate
$V_{\mathrm{sem}}(t,c)$ requires that the rendered prompt preserve the canonical
facts, operations, relations, and answer equivalence class of the original
task. The online runner samples only from the structurally feasible set
\[
    \Omega_{\mathrm{auto}}
    =
    \{(t,c): t\in\mathcal{T}, c\in\mathcal{C},
    V_{\mathrm{auto}}(t,c)=1\},
\]
while confirmed model-error claims are restricted to candidates that also
satisfy $V_{\mathrm{sem}}(t,c)=1$ and pass response-level adjudication.

\paragraph{Automatic routing signal and audited failure label.}
Let $M$ be a fixed model interface and let $E$ be an answer extractor that maps
a raw model response to a predicted answer $\hat a$. For a structurally valid
candidate $x=(t,c,R(t,c))$, the automatic mismatch indicator is
\[
    F_M(x) =
    \mathbb{1}\!\left[
    \operatorname{norm}(E(M(R(t,c)))) \ne \operatorname{norm}(a_t)
    \right],
\]
where $\operatorname{norm}$ handles exact answers, yes/no answers with trailing
explanation, and single-integer answers embedded in a sentence. This quantity is
a routing signal: it identifies candidates for later review, but it is not
itself a scientific failure label.

We reserve $A_M(x)=1$ for a resolved audited model error. This label is
assigned only when the rendered prompt is semantically valid, the model's final
answer is identifiable from the raw response, the extractor has not introduced a
formatting artifact, and the normalized final answer is not equivalent to
$a_t$. Candidates without this resolved label are not counted as confirmed
model errors, regardless of their automatic mismatch status.

\paragraph{Construction policies.}
The experiments compare two policies over the same task bank and grammar.
Uniform component sampling is the equal-budget control: it samples each
component option with equal probability. CAPS uses the same budget, renderer,
model interface, and answer extractor, but adapts component probabilities using
observed mismatch signals.

CAPS maintains a real-valued score table
$s=\{s_{g,j}: g \in G, j \in \mathcal{C}_g\}$ over component options. These
scores are internal run state only; they are never submitted to the model. At
iteration $i$, for group $g$ and option $j$, the CAPS sampling probability is
\[
    \pi_i(c^g=j)
    =
    \frac{\max(\epsilon,\, 1 + s_{g,j}/\tau_i)}
    {\sum_{k \in \mathcal{C}_g}
    \max(\epsilon,\, 1 + s_{g,k}/\tau_i)},
    \qquad \epsilon=10^{-3}.
\]
The temperature schedule is
\[
    \tau_i = \max(0.2, 1 - i/I),
\]
where $I$ is the number of iterations. A component assignment is drawn by
sampling one option from each group under $\pi_i$ and then rendering the
resulting prompt. For each evaluated candidate with routing signal $r=F_M(x)$,
only the selected component options receive an exponential-moving update:
\[
    s_{g,c^g}
    \leftarrow
    s_{g,c^g} + \eta\left(r - s_{g,c^g}\right),
    \qquad \eta=0.05.
\]
Unselected options are unchanged. Components repeatedly observed in mismatched
prompts therefore become more likely to be sampled in later batches, while the
temperature floor preserves exploration. In the neutral-prior setting all
component scores are initialized to zero; in earlier diagnostic configurations,
fixed stressor priors are declared in the config and snapshotted with the run.
In all settings, task identifiers are sampled uniformly with replacement; only
prompt-component options are adapted.

\paragraph{Run protocol and artifacts.}
The unit of construction is a batch rather than an isolated prompt. Let
$B_i=\{x_{i,b}\}_{b=1}^{m}$ be the batch at iteration $i$, and let
$A=\bigcup_{i=1}^{I}B_i$ be the queried candidate set under total budget
$Q=Im$. Each run first initializes a state consisting of the
configuration, task-bank snapshot, seed, component scores, query budget,
construction policy, and model adapter. Given this state, the runner executes a
matched construction loop: it samples tasks with replacement, samples component
assignments, renders prompts, checks structural validity, calls the adapter,
extracts answers, and updates CAPS scores when CAPS is active.

After every iteration, the run appends a trajectory record and refreshes a
status record. The trajectory stores the evaluated batch, the iteration-best
candidate, the temperature, the policy name, and the score table. The status
record stores the best candidate seen so far together with resume metadata.
Every candidate record stores the task identifier, task family, canonical
answer, selected components, rendered prompt, validity errors, raw model
response, extracted answer, correctness flag, and failure score. These artifacts
make each run durable, resumable, and auditable.

\paragraph{Audit protocol and endpoints.}
Automatic checks catch malformed artifacts but do not establish semantic
equivalence. The audit therefore separates three decisions. First, prompt-level
semantic validity asks whether the rendered prompt preserves the task's
entities, relations, operations, and answer equivalence class. Second,
extraction validity asks whether the stored extracted answer reflects the
model's final answer rather than a parsing or formatting artifact. Third,
model-error adjudication assigns $A_M(x)=1$ only when the prompt is
semantically valid, extraction is valid, and the model's final answer is not
equivalent to the canonical answer. Candidates that fail any criterion, remain
unaudited, or stay disputed after review are excluded from audited failure
yield.

Our primary endpoint is audited yield over the matched query budget,
\[
    Y(A)
    =
    \frac{1}{Q}
    \sum_{x\in A}
    A_M(x).
\]
Unaudited or rejected candidates contribute zero. This makes $Y(A)$ a
conservative endpoint whose positive counts come only from resolved audited
failures. We also report query-to-first-audited-failure,
\[
    T_1 = \min\{q: A_M(x_q)=1\},
\]
when such a candidate exists. In addition, we distinguish raw proxy mismatch
counts from audited outcomes, and record-level audited counts from unique
audited prompt keys, so that repeated renderings of the same task-component
assignment are not treated as independent discoveries. With a binary routing
signal, best-score comparisons saturate quickly; a policy advantage would need
to appear in audited yield, first-failure time, or unique audited prompt keys.

\paragraph{Matched comparison.}
CAPS and uniform component sampling share the same task bank, renderer, model
adapter, answer extractor, validity checks, iteration count, batch size, seed
list, and record format. The only operational difference is how prompt
components are sampled. This keeps the comparison deliberately narrow while
holding query budget and the rest of the evaluation pipeline fixed. Within each
seed and iteration,
task draws follow the same deterministic stream; component choices are not
paired because the component policy is the object under comparison. The
neutral-prior configuration is therefore a descriptive zero-prior check rather
than a fully isolating causal test of the CAPS update rule, and none of the
comparisons expands the prompt space or uses extra model calls.

\paragraph{Method boundary.}
The contribution of the method is the decomposition of targeted test
construction into four auditable objects: a frozen task bank, a finite
perturbation grammar, matched sampling policies over that grammar, and a
checkpointed artifact-and-audit contract. This makes sampled candidates
reconstructable and reviewable from recorded artifacts. The method does not
provide global prompt-space coverage, an exact replay guarantee for every
execution environment, or a replacement for broad benchmark evaluation. Its
intended role is narrower: to produce compact, inspectable stress tests that
complement static reasoning benchmarks.

\section{Experiments}
We evaluate targeted reasoning tests through matched comparisons rather than
through isolated probe runs. The experiments ask three questions: whether the
artifact contract is reproducible, whether the targeted banks contain prompt
variants that survive audit as model-error evidence, and whether CAPS improves
audited discovery over equal-budget uniform component sampling. Throughout, we
keep proxy mismatch signals separate from audited failure labels. Automatic
mismatches are used only to route candidates for review; model-error claims are
made only from resolved audit outcomes.

\paragraph{Targeted probe banks and matched setup.}
The initial v1/v2 probe bank contains 12 hand-authored reasoning tasks drawn
from three families: arithmetic problems with tempting numerical shortcuts,
symbolic-ordering or relation-tracking tasks, and short abstract-rule tasks.
The v3 bank expands the same design to 18 tasks while preserving the same
prompt-component grammar. In the results below, v1 and v2 denote 12-task bank
revisions, v3 denotes the 18-task bank, and v4 denotes a neutral-prior follow-up
on the v3 bank rather than a new task bank. These banks are targeted probes,
not broad reasoning benchmarks: they are designed to expose brittle behavior
under controlled prompt perturbations while keeping semantic review feasible.

All CAPS-versus-uniform comparisons use the same task bank, renderer, answer
extractor, query budget, and audit path. The v1--v3 model-facing comparisons
use three seeds, five iterations, and batch size three, yielding 90 preserved
responses per matched comparison. The v4 neutral-prior follow-up uses four
seeds, eight iterations, and batch size three, yielding 192 preserved
responses. Model-facing runs use fixed local Hugging Face models in the same
GPU-backed environment and preserve raw responses and extracted answers so that
all proxy mismatches can be reviewed under the same audit contract.

\paragraph{Artifact-level validation.}
Before turning to model behavior, we validate the comparison object itself with
a deterministic fixture adapter that uses the same record format as the
model-facing runs. Under this controlled fixture, CAPS and uniform component
sampling both reach a mean best score of 0.8115 over five seeds, producing five
ties (Table~\ref{tab:synthetic_comparison_summary}). This result is not evidence
about language-model reasoning. It verifies that task-set loading,
checkpointing, baseline comparison, and table export behave as intended before
mismatch signals are supplied by real model responses.

\paragraph{Proxy mismatch results.}
Tables~\ref{tab:a100_v2_model_facing_summary},
\ref{tab:a100_v3_model_facing_summary}, and
\ref{tab:a100_v4_model_facing_summary} report automatic proxy mismatches before
final audit. These tables show that the probe banks do surface brittle prompt
variants, but they do not support a strong claim for CAPS. Across the completed
matched runs, best-score ties are common, and uniform component sampling often
finds as many proxy mismatches as CAPS. In the v3 Qwen slice, for example,
uniform records one seed-pair win and two ties. The v4 neutral-prior follow-up
shows the same pattern more clearly: it yields 26 proxy mismatches, with 14
under CAPS and 12 under uniform sampling, but all four seed-pair best-score
comparisons are ties and 12 of the 13 unique mismatch prompt keys are shared by
both strategies. The proxy evidence therefore shows that both policies can
enter the same small brittle region of the prompt-component space; it does not
by itself, establish a discovery advantage for CAPS.

\paragraph{Audited results.}
The audit tables convert routed proxy mismatches into scientific claims by
requiring semantic validity, extraction validity, and answer non-equivalence.
For v2, resolved second review confirms 32 model-error rows and excludes one
extraction artifact (Table~\ref{tab:a100_v2_resolved_audit_summary}). These
confirmed rows span arithmetic, symbolic-relation, and counterfactual tasks,
showing that the targeted bank does contain reviewable model-facing failures.
However, they do not favor CAPS: Qwen has more confirmed uniform-sampling rows
than CAPS rows, Phi-3 is tied, and Mistral has one more confirmed uniform row
after the excluded CAPS extraction artifact.

For v3, resolved second review confirms five model-error rows across three
unique prompt keys and excludes one answer-format artifact
(Table~\ref{tab:a100_v3_resolved_audit_summary}). The confirmed failures include
gate-counterfactual, ticket-count arithmetic, and key-containment relation
cases. This strengthens the claim that the expanded bank contains audited
prompt-level failures, but again does not show a matched-budget advantage for
CAPS.

The v4 neutral-prior follow-up provides a stronger prompt-key analysis. After
deduplicating 26 proxy mismatches into 13 prompt keys, resolved review confirms
12 unique prompt keys corresponding to 24 preserved response rows and excludes
one yes/no formatting artifact
(Table~\ref{tab:a100_v4_resolved_audit_summary}). Because review is keyed by
prompt identity before row expansion, the unique prompt-key count is the primary
discovery unit and the 24 rows are secondary evidence. As in v2 and v3, the
targeted bank contains audited failures, but the neutral-prior setting still
does not show higher audited yield or unique-key discovery for CAPS than for
matched uniform sampling.

Taken together, the audited results support two conclusions. First, small
targeted probe banks can produce model-facing prompt variants that survive
semantic and extraction review as confirmed errors. Second, under the matched
budgets studied here, CAPS does not outperform uniform component sampling on the
endpoints that matter for this paper: audited yield, query-to-first-audited
failure, or unique audited prompt keys. Table~\ref{tab:baseline_separation_summary}
consolidates this negative matched-baseline result while keeping row-level
audited counts separate from unique prompt-key discovery.

\begin{table}[!htbp]
\centering
\scriptsize
\caption{Audited baseline separation for model-facing targeted tests. The table counts only resolved rows labeled as confirmed model errors; excluded answer-format or extraction artifacts are omitted from the confirmed-row columns. The unique-key column separates shared and strategy-only discoveries. The rightmost column is descriptive and does not turn CAPS into the paper's main contribution.}
\label{tab:baseline_separation_summary}
\resizebox{\linewidth}{!}{%
\begin{tabular}{@{}lllll@{}}
\toprule
Evidence slice & Budget & Confirmed rows & Unique keys & Audit-safe reading \\
\midrule
v2, three models & 270 responses & 15 CAPS / 17 Uniform & 9 shared / 1 CAPS-only / 3 Uniform-only & Uniform matches or exceeds CAPS under the audited rows. \\
v3, Qwen & 90 responses & 2 CAPS / 3 Uniform & 2 shared / 0 CAPS-only / 1 Uniform-only & Uniform has the larger audited count in this slice. \\
v4 neutral, Qwen & 192 responses & 13 CAPS / 11 Uniform & 11 shared / 1 CAPS-only / 0 Uniform-only & Mostly shared keys; no superiority claim follows. \\
\bottomrule
\end{tabular}
}
\end{table}

\section{Discussion}
The main empirical finding is limited but straightforward. CAPS and uniform
component sampling both surface prompt-conditioned errors that survive audit,
while the completed matched budgets do not show that CAPS discovers more
audited failures or more unique audited prompt keys. In a small structured
prompt-component grammar, uniform component sampling can already enter brittle
regions. As a result, targeted-test studies should not rely on proxy mismatch
counts, selected examples, or unmatched sampling policies.

The protocol raises the standard for what counts as evidence. An answer-key
mismatch is not enough; a candidate must also survive semantic review and
extraction review. This makes the resulting evidence more falsifiable than
ordinary example-driven reports because it permits checking whether the prompt
preserved the underlying task, whether extraction altered the model's answer,
whether repeated rows correspond to the same prompt key, and whether the same
query budget was spent by each construction policy.

The negative result also clarifies the role of CAPS. CAPS is a candidate
proxy-guided sampler, not the central scientific claim. Its purpose in this
paper is to test a natural hypothesis: whether reusing automatic mismatch
signals to bias later component choices improves audited discovery over uniform
component sampling. The present answer is no under the tested budgets and task
banks. That answer makes equal-budget uniform component sampling part of
the methodological contribution rather than a disposable control. Future
adaptive policies should be evaluated against the same audit contract and should
claim an advantage only when gains appear in audited yield, unique audited
prompt keys, or query-to-first-audited-failure, not only in proxy signals.

\section{Limitations}
The targeted-test protocol is an evaluation artifact, not a general solution to
reasoning evaluation.

\paragraph{Task coverage.}
The versioned task banks contain 12 to 18 hand-authored seed tasks across
arithmetic, symbolic relations, and short counterfactual/default-rule cases. The
task-bank cards provide derivation checks and perturbation-validity notes, not
independent task-bank certification. This scale is sufficient for artifact
validation and compact model-facing probes, but it is not a broad reasoning
benchmark. Results should be interpreted as targeted stress-test evidence for
the specified task families and prompt components, not as coverage of reasoning
behavior in general.

\paragraph{Controlled artifact evidence.}
The fixture CAPS-vs-uniform comparison verifies the artifact pipeline; it is not
evidence about any deployed language model. The A100-backed Qwen2.5-7B,
Phi-3 Mini, and Mistral-7B roots are model-facing evidence, but their task-bank
versions and audit status must remain separated. v1 roots are proxy-only triage
evidence and are retained only as artifact-contract checks. The audited v2, v3,
and neutral-prior v4 roots support targeted candidate-failure claims within
their probe-bank versions, but they do not support broad reasoning-coverage
claims or a CAPS advantage over uniform component sampling. Stronger
model-facing claims would require larger frozen task banks, repeated runs across
models and seeds, and broader manual review of proxy-mismatched prompt variants.

\paragraph{Prompt-space coverage.}
The protocol samples from a constrained prompt-component space. It does not
enumerate all prompts that could test a task, and it does not prove that
discovered prompts are globally strongest stress tests. CAPS and uniform
component sampling are tools for producing auditable targeted tests, not
complete prompt-space search procedures.

\paragraph{Validity and scoring.}
The validity checks are structural: they enforce prompt length, task
anchors, answer-key presence, and stored validity tags. They do not prove that a
prompt preserves every semantic detail of the original task after perturbation.
For real model runs, proxy-mismatched candidates should be manually audited before
being counted as failures. Answer extraction is also simple: the adapters read a
final answer line, or fall back to a conservative normalized-answer search, and
compare the result against a canonical answer with simple normalization rules.
This can undercount partially correct reasoning, overcount formatting errors,
and miss cases where the model gives the right answer for the wrong reason. The
binary failure score also makes best-score comparisons insensitive once both
strategies find at least one error, so policy claims should rely on audited-yield
and unique-key metrics rather than best-score ties alone.

\paragraph{Execution reproducibility.}
GPU-backed runs use spot capacity when available. Spot eviction can interrupt a
run, and endpoint availability can change between scheduled heartbeats. These
risks are mitigated by checkpointing and by recording model revisions and run
metadata, but they are not eliminated. A complete rerun still depends on the
recorded endpoint or model revision together with the saved configuration and
task-bank state. Later v3/v4 runs provide stronger revision-level provenance
than the earlier v2 runs, whose snapshots support the reported audited rows but
not equally strong commit-level rerun provenance.

\section{Conclusion}
This paper studies targeted prompt variation as an auditable evaluation
object. It asks whether a small, controlled sampling process over prompt
components with documented semantic-validity criteria can produce inspectable
stress tests under a fixed query budget. The key design choice is to separate three objects
that are often conflated: an automatic proxy mismatch used as a
candidate-routing signal, a structurally valid rendered prompt, and an audited
model-error claim.

The completed study shows both the value and the limit of this formulation. The
audits confirm model-error rows for targeted prompts in arithmetic,
relation-tracking, and counterfactual/default-rule tasks. The primary discovery
unit is the audited prompt key; record-level rows are budget and traceability
evidence and are not independent prompt discoveries when the same
task-component assignment repeats. At the same time, the completed matched
budgets, including the neutral-prior follow-up, do not establish higher audited
yield or unique-key discovery for CAPS than for matched uniform component
sampling. The protocol can therefore produce inspectable evidence about
prompt-conditioned failures, but the proxy-guided sampler does not outperform
the matched uniform control in the current study. Future work should scale the
task bank and improve sampling policies only under the same audit contract that
keeps the present claims narrow.

\bibliographystyle{plainnat}
\bibliography{references}

\clearpage
\appendix
\section{Appendix}
This appendix records the audit rules and artifact pointers needed to interpret
the targeted-test evidence without treating automatic mismatches as model-error
claims.

The seed task bank is documented in \texttt{docs/task\_bank\_card.md}. The card
lists each task identifier, gold answer, answer derivation, perturbation-validity
checklist, prompt-grammar coverage, and audit criteria. This appendix does not
duplicate the full card so the machine-readable task set and the audit artifact
remain the source of record.

\paragraph{Artifact traceability.}
The model-facing evidence has a project-local traceability runbook under
\texttt{docs/}. Each paper-level claim must trace through a result root,
per-seed \texttt{trajectory.jsonl} files with raw responses, analysis exports,
resolved audit rows when failures are claimed, and the corresponding paper
table. The current audited claims use the v2, v3, and neutral-prior v4 final
audit summaries; v1 roots remain proxy-only or triage evidence. This rule is
also why the paper reports proxy mismatches, audited-yield numerators, unique
prompt keys, and best-score diagnostics as distinct quantities.

\paragraph{Audit rubric.}
Table~\ref{tab:audit_rubric} summarizes the decision rules used to convert
automatic mismatches into model-error evidence. The rubric is deliberately
stricter than the automatic answer-key comparison: a mismatch can route a
candidate to review, but it cannot by itself support a scientific failure
claim.

Table~\ref{tab:evaluation_protocol} summarizes how the paper maps reported
quantities to evidence artifacts. It is included here because it clarifies the
relationship between proxy signals, audited outcomes, and supporting records,
but it is supplementary to the main experimental comparison.

\begin{table}[!htbp]
\centering
\scriptsize
\caption{Audit rubric for resolving automatic mismatches into paper-level
model-error evidence.}
\label{tab:audit_rubric}
\begin{tabular}{p{0.22\linewidth}p{0.68\linewidth}}
\toprule
Decision & Rule \\
\midrule
Prompt semantic validity & The rendered prompt must preserve the task's
entities, relations, operations, constraints, and answer equivalence class. A
prompt that changes the task is excluded even if the model answer disagrees
with the stored answer key. \\
Extraction validity & The extracted answer must reflect the model's final
answer in the preserved raw response. Rows caused by parser behavior,
formatting ambiguity, or yes/no answer-format artifacts are excluded. \\
Confirmed model-error label & A row receives \texttt{confirmed\_model\_error}
only when the prompt is semantically valid, extraction is valid, adjudication is
resolved, and the model's final answer is not equivalent to the canonical
answer. \\
Review unit & v2 and v3 audits use row-level second review of proxy mismatches
with the preliminary triage label hidden from the reviewer; strategy and model
fields remain present for traceability. The v4 neutral-prior audit deduplicates
proxy mismatches by prompt key, reviews those prompt keys, and expands resolved
labels back to matching response rows. \\
Row and key accounting & Record-level confirmed rows are reported separately
from unique audited prompt keys. Repeated rows with the same task-component
assignment are evidence of repeated observations, not independent prompt
discoveries. \\
\bottomrule
\end{tabular}
\end{table}

\begin{table}[!htbp]
\centering
\scriptsize
\caption{Estimand-to-artifact map for targeted reasoning tests. The table
separates infrastructure checks, candidate-routing proxy signals, and resolved
audit labels so that automatic mismatches are not read as scientific failure
claims.}
\label{tab:evaluation_protocol}
\begin{tabular}{p{0.24\linewidth}p{0.42\linewidth}p{0.24\linewidth}}
\toprule
Reported quantity & Interpretation & Evidence artifact \\
\midrule
Artifact reproducibility & Same task bank, renderer, budget, checkpoint schema,
and analysis path across strategies & config, task-set, trajectory, and status
snapshots \\
Proxy mismatch count & Automatic answer-key disagreement before semantic and
extraction audit; candidate-routing signal only & model-facing summaries and raw
trajectories \\
Audited-yield numerator & Confirmed record-level model-error rows, the
unnormalized numerator of $Y(A)$ under fixed matched budgets & resolved audit
tables and audit-row files \\
Unique prompt keys & Distinct task-component assignments after deduplication;
repeated rows are not independent discoveries & resolved audit tables and
audit manifests \\
Matched best score & Binary saturation diagnostic for whether a strategy found
at least one proxy-mismatched candidate; not discovery efficiency by itself &
comparison summaries \\
\bottomrule
\end{tabular}
\end{table}

\paragraph{Proxy-only v1 artifact check.}
Table~\ref{tab:a100_model_facing_summary} records the v1 A100 proxy-only runs.
It is retained as an artifact-contract check and is not used for audited
failure-yield claims because its single-review triage labels were never resolved
through the paper's final audit protocol.

\begin{table}[!htbp]
\centering
\small
\caption{Proxy-only A100-backed runs on \texttt{reasoning\_probe\_v1}. Proxy mismatches are automatic answer-mismatch signals; preliminary triage rows are single-review labels, not completed audit evidence or failure-yield estimates. All rows use the same v1 12-task probe bank, three seeds, five iterations, and batch size three per strategy. Audited v2--v4 roots are reported separately.}
\label{tab:a100_model_facing_summary}
\begin{tabular}{llrrrr}
\toprule
Model & Strategy & Candidates & Proxy mismatches & Mismatch rate & Prelim. triage rows \\
\midrule
Qwen2.5-7B & CAPS & 45 & 2 & 0.0444 & 2 \\
Qwen2.5-7B & Uniform & 45 & 3 & 0.0667 & 3 \\
Phi-3 Mini & CAPS & 45 & 2 & 0.0444 & 2 \\
Phi-3 Mini & Uniform & 45 & 2 & 0.0444 & 2 \\
Mistral-7B & CAPS & 45 & 10 & 0.2222 & 8 \\
Mistral-7B & Uniform & 45 & 10 & 0.2222 & 9 \\
\bottomrule
\end{tabular}
\end{table}

\paragraph{Detailed model-facing tables.}
The main text reports the consolidated audited baseline-separation table. The
tables below retain the detailed fixture, proxy, and resolved-audit summaries
used to support those counts. They are grouped here before the runtime and
release metadata so proxy and resolved-audit counts can be read together.

\begin{table}[t]
\centering
\small
\caption{Synthetic CAPS-vs-uniform smoke comparison. This table validates the analysis artifact contract and is not a model-facing result.}
\label{tab:synthetic_comparison_summary}
\begin{tabular}{lr}
\toprule
Metric & Value \\
\midrule
Seeds & 5 \\
CAPS mean best score & 0.8115 \\
Uniform mean best score & 0.8115 \\
Mean best-score delta & 0.0 \\
CAPS wins & 0 \\
Uniform wins & 0 \\
Ties & 5 \\
CAPS win rate & 0.0 \\
\bottomrule
\end{tabular}
\end{table}

\begin{table}[t]
\centering
\small
\caption{A100-backed follow-up runs on \texttt{reasoning\_probe\_v2}. Proxy mismatches are automatic answer-mismatch signals. Unique keys are counted within each strategy row. Triage model-error rows are preliminary single-review labels and exclude rows marked as answer-format artifacts; they are not audited failure-yield estimates.}
\label{tab:a100_v2_model_facing_summary}
\begin{tabular}{llrrrr}
\toprule
Model & Strategy & Candidates & Proxy mismatches & Strategy keys & Triage rows \\
\midrule
Qwen2.5-7B & CAPS & 45 & 3 & 3 & 3 \\
Qwen2.5-7B & Uniform & 45 & 4 & 4 & 4 \\
Phi-3 Mini & CAPS & 45 & 2 & 2 & 2 \\
Phi-3 Mini & Uniform & 45 & 2 & 2 & 2 \\
Mistral-7B & CAPS & 45 & 11 & 10 & 10 \\
Mistral-7B & Uniform & 45 & 11 & 10 & 11 \\
\bottomrule
\end{tabular}
\end{table}

\begin{table}[t]
\centering
\small
\caption{Resolved second-review status for \texttt{reasoning\_probe\_v2} proxy mismatches. Counts are record-level within each model--strategy row; unique keys are deduplicated only within that row, and repeated keys across rows, strategies, or models are not independent discoveries. Confirmed errors include only resolved rows; excluded rows are not counted as audited failures.}
\label{tab:a100_v2_resolved_audit_summary}
\begin{tabular}{llrrrrr}
\toprule
Model & Strategy & Rows & Reviewed & Confirmed errors & Excluded & Unique keys \\
\midrule
Qwen2.5-7B & CAPS & 3 & 3 & 3 & 0 & 3 \\
Qwen2.5-7B & Uniform & 4 & 4 & 4 & 0 & 4 \\
Phi-3 Mini & CAPS & 2 & 2 & 2 & 0 & 2 \\
Phi-3 Mini & Uniform & 2 & 2 & 2 & 0 & 2 \\
Mistral-7B & CAPS & 11 & 11 & 10 & 1 & 9 \\
Mistral-7B & Uniform & 11 & 11 & 11 & 0 & 10 \\
\bottomrule
\end{tabular}
\end{table}

\begin{table}[t]
\centering
\small
\caption{Qwen2.5-7B-Instruct comparison on the 18-task \texttt{reasoning\_probe\_v3} bank. Record, proxy-mismatch, and audited-error counts are record-level over preserved A100 responses. The best-score mean is computed over the three matched seed-pair runs and is reported only as a small-budget proxy-saturation diagnostic.}
\label{tab:a100_v3_model_facing_summary}
\begin{tabular}{lrrrr}
\toprule
Strategy & Records & Proxy mismatches & Audited model-error rows & Mean best score over 3 seeds \\
\midrule
CAPS & 45 & 3 & 2 & 0.3333 \\
Uniform & 45 & 3 & 3 & 0.6667 \\
\bottomrule
\end{tabular}
\end{table}

\begin{table}[t]
\centering
\small
\caption{Resolved second-review status for the Qwen2.5-7B-Instruct \texttt{reasoning\_probe\_v3} proxy mismatches. Counts are record-level; repeated prompt keys across strategies are not independent discoveries. Confirmed errors include only resolved rows; excluded rows are not counted as audited failures.}
\label{tab:a100_v3_resolved_audit_summary}
\begin{tabular}{llrrrrr}
\toprule
Model & Strategy & Rows & Reviewed & Confirmed errors & Excluded & Unique keys \\
\midrule
Qwen2.5-7B & CAPS & 3 & 3 & 2 & 1 & 2 \\
Qwen2.5-7B & Uniform & 3 & 3 & 3 & 0 & 3 \\
\bottomrule
\end{tabular}
\end{table}

\begin{table}[t]
\centering
\scriptsize
\caption{Qwen2.5-7B-Instruct neutral-prior v4 comparison on the 18-task
\texttt{reasoning\_probe\_v3} bank. Proxy mismatches are automatic
answer-mismatch signals from preserved A100 responses; audited errors include
only resolved rows from prompt-key review. All four seed pairs tie on best
score, so the row difference is descriptive rather than evidence of a
CAPS advantage; within-strategy key counts are not independent
discoveries after overlap.}
\label{tab:a100_v4_model_facing_summary}
\begin{tabular}{lrrrrr}
\toprule
Strategy & Records & Proxy & Audited errors & Within-strategy keys & Mean best \\
\midrule
CAPS & 96 & 14 & 13 & 12 & 1.0000 \\
Uniform & 96 & 12 & 11 & 11 & 1.0000 \\
\bottomrule
\end{tabular}
\end{table}

\begin{table}[t]
\centering
\small
\caption{Resolved audit status for the Qwen2.5-7B-Instruct neutral-prior v4 proxy mismatches. Counts are record-level over preserved A100 responses; repeated prompt keys across strategies are not independent discoveries. Confirmed errors include only resolved rows; excluded rows are not counted as audited failures.}
\label{tab:a100_v4_resolved_audit_summary}
\begin{tabular}{llrrrrr}
\toprule
Model & Strategy & Rows & Reviewed & Confirmed errors & Excluded & Unique keys \\
\midrule
Qwen2.5-7B & CAPS & 14 & 14 & 13 & 1 & 12 \\
Qwen2.5-7B & Uniform & 12 & 12 & 11 & 1 & 11 \\
\bottomrule
\end{tabular}
\end{table}

\clearpage

\paragraph{Model and runtime metadata.}
The model-facing roots were executed on an Azure East Japan A100-class spot VM
with \texttt{torch==2.6.0+cu126} and a local Hugging Face adapter. Generation
uses \texttt{max\_new\_tokens=192}; the v3
and v4 reproducibility captures record \texttt{bfloat16},
\texttt{device\_map=auto}, Python 3.11.15, \texttt{uv 0.11.8}, CUDA 12.6
packages, and an A100 80GB GPU.
The audited claims use these model identifiers: v2 uses
\path{Qwen/Qwen2.5-7B-Instruct},
\path{microsoft/Phi-3-mini-4k-instruct}, and
\path{mistralai/Mistral-7B-Instruct-v0.3}; their result snapshots do not record
resolved Hugging Face commit SHAs. The later v3 and v4 Qwen runs record and pin
the resolved model revision, respectively. The supplemental
artifact records current license-verification status in
\path{docs/asset_license_status.md} and exact revision-evidence boundaries in
\path{docs/model_revision_evidence.md}; compact Hugging Face metadata snapshots
verify the model license fields, while broader release questions about
model-card terms and preserved responses are tracked separately.

\paragraph{Artifact manifest.}
Table~\ref{tab:artifact_manifest} gives the compact manifest for the audited
evidence used in the paper. All result roots listed in the table are under
\path{experiments/results_hf_local/}; all resolved audit files are under
\path{experiments/audits/}. The raw-response evidence for every model-facing
claim is the corresponding \path{runs/*/trajectory.jsonl} file under each result
root.
The public release policy in \path{docs/release_packaging_policy.md} separates
this local evidence archive from a supplemental package: raw model responses are
not included in the recommended public package until model-card response
redistribution terms are reviewed. The default public package therefore uses
sanitized public audit rows with prompt text and response hashes, final audit
summaries, configs, analysis summaries, and compact metadata; local final audit
rows or second-review templates containing raw response text remain controlled
evidence outside the default public package. The default public package supports
table and count reproduction from sanitized rows, summaries, and hashes; exact
raw-response reinspection requires the controlled local archive or rerunning the
released configs against the recorded model identifiers, and against recorded
revisions where revision evidence is available. For v2, the rerun provenance is
identifier-level rather than commit-level; v3 and v4 record the resolved or
pinned Qwen revision.

\begingroup
\footnotesize
\sloppy
The exact audited result-root suffixes are:
\begin{itemize}[leftmargin=*]
    \item v2 Qwen: \path{20260502T082056Z-qwen25-7b-v2-seeds23-29-31}.
    \item v2 Phi-3: \path{20260502T080133Z-phi3-mini-v2-seeds23-29-31}.
    \item v2 Mistral: \path{20260502T084226Z-mistral7b-v2-seeds23-29-31}.
    \item v3 Qwen: \path{20260502T104746Z-qwen25-7b-v3-seeds23-29-31}.
    \item v4 neutral-prior Qwen:
    \path{20260502T114847Z-qwen25-7b-neutral-v4-seeds23-29-31-37}.
\end{itemize}
\endgroup

\begin{table}[H]
\centering
\scriptsize
\caption{Audited evidence manifest. Root suffixes are relative to the
project-local Hugging Face result directory.}
\label{tab:artifact_manifest}
\begin{tabular}{p{0.17\linewidth}p{0.40\linewidth}p{0.33\linewidth}}
\toprule
Claim surface & Result roots & Required audit and table artifacts \\
\midrule
v2 audited rows &
Qwen v2; Phi-3 v2; Mistral v2 &
sanitized public audit rows plus final audit summary; v2 model-facing and
resolved-audit paper tables \\
v3 audited Qwen rows &
Qwen v3 &
sanitized public audit rows plus final audit summary; v3 model-facing and
resolved-audit paper tables \\
v4 neutral-prior audited Qwen rows &
Qwen v4 neutral-prior &
sanitized public audit rows, final audit summary, and second-review manifest;
v4 model-facing and resolved-audit paper tables \\
\bottomrule
\end{tabular}
\end{table}

\clearpage

\paragraph{Responsible use and release.}
The artifact is intended for diagnostic evaluation and reproducibility, not for
ranking models or certifying broad reasoning ability. The main positive use is
to make targeted failure evidence reviewable from preserved prompts, raw
responses, and audit rows. The main misuse risk is overclaiming from a small
targeted probe bank; the protocol mitigates this by separating fixture outputs,
proxy mismatches, resolved model-error rows, and unique prompt keys, and by
requiring narrow task-bank and prompt-grammar scope statements.

\section*{NeurIPS Paper Checklist}

\begin{enumerate}

\item {\bf Claims}
    \item[] Question: Do the main claims made in the abstract and introduction accurately reflect the paper's contributions and scope?
    \item[] Answer: \answerYes{}.
    \item[] Justification: The abstract and introduction state the protocol-first contribution, the narrow targeted-probe scope, and the negative CAPS-versus-uniform finding. The limitations section further restricts the claims to the audited task banks and prompt-component grammar.

\item {\bf Limitations}
    \item[] Question: Does the paper discuss the limitations of the work performed by the authors?
    \item[] Answer: \answerYes{}.
    \item[] Justification: The paper includes a dedicated Limitations section covering task coverage, fixture-only evidence, prompt-space coverage, validity and extraction risks, and spot-VM execution reproducibility.

\item {\bf Theory assumptions and proofs}
    \item[] Question: For each theoretical result, does the paper provide the full set of assumptions and a complete (and correct) proof?
    \item[] Answer: \answerNA{}.
    \item[] Justification: The paper does not present theoretical results, theorems, or formal proofs; it presents an evaluation protocol and audited empirical artifact.

    \item {\bf Experimental result reproducibility}
    \item[] Question: Does the paper fully disclose all the information needed to reproduce the main experimental results of the paper to the extent that it affects the main claims and/or conclusions of the paper (regardless of whether the code and data are provided or not)?
    \item[] Answer: \answerYes{}.
    \item[] Justification: The method, experiments, appendix, and supplemental project files specify the task banks, prompt grammar, seeds, budgets, configs, checkpoint schema, audit rows, and analysis tables used for the main claims. The local evidence archive preserves raw-response trajectories; the default public supplement releases sanitized audit rows, prompt hashes, raw-response hashes, and response lengths rather than raw response text.

\item {\bf Open access to data and code}
    \item[] Question: Does the paper provide open access to the data and code, with sufficient instructions to faithfully reproduce the main experimental results, as described in supplemental material?
    \item[] Answer: \answerNo{}.
    \item[] Justification: The supplemental package boundary is specified by a generated manifest and release policy. The public package includes task banks, configs, scripts, tests, audit summaries, sanitized public audit rows, result tables, and README commands needed to reproduce the reported tables and claim checks, while excluding secrets and raw model-response text by default. Exact response-text reinspection requires either the controlled local evidence archive or rerunning the recorded configs when the model and runtime remain available. The answer is therefore conservative: code and sanitized audit artifacts are provided, but the default public package is not a full raw-response data release. The paper does not require releasing model weights.

\item {\bf Experimental setting/details}
    \item[] Question: Does the paper specify all the training and test details (e.g., data splits, hyperparameters, how they were chosen, type of optimizer) necessary to understand the results?
    \item[] Answer: \answerYes{}.
    \item[] Justification: There is no training procedure. The experiments section and appendix specify the tested models, task-bank revisions, seeds, iterations, batch sizes, construction policies, neutral-prior follow-up, adapter role, and audit procedure.

\item {\bf Experiment statistical significance}
    \item[] Question: Does the paper report error bars suitably and correctly defined or other appropriate information about the statistical significance of the experiments?
    \item[] Answer: \answerNo{}.
    \item[] Justification: The study reports audited counts, matched seed-pair summaries, and unique prompt-key counts rather than formal confidence intervals. The paper explicitly treats the results as a compact diagnostic study and does not claim statistically significant CAPS superiority.

\item {\bf Experiments compute resources}
    \item[] Question: For each experiment, does the paper provide sufficient information on the computer resources (type of compute workers, memory, time of execution) needed to reproduce the experiments?
    \item[] Answer: \answerYes{}.
    \item[] Justification: The model-facing runs are identified as Azure East Japan A100-class runs, and public runtime metadata records the checkpointing contract, CUDA/PyTorch environment, and exact resolved or pinned Qwen revisions for v3/v4 roots. The v2 roots are explicitly marked as weaker identifier-only provenance. The operational cloud runbook is kept outside the default public package.
    
\item {\bf Code of ethics}
    \item[] Question: Does the research conducted in the paper conform, in every respect, with the NeurIPS Code of Ethics \url{https://neurips.cc/public/EthicsGuidelines}?
    \item[] Answer: \answerYes{}.
    \item[] Justification: The work uses hand-authored synthetic reasoning tasks and evaluates existing model interfaces without collecting personal data, involving human subjects, or deploying a user-facing system. The artifact preserves evidence for audit while avoiding secret disclosure.

\item {\bf Broader impacts}
    \item[] Question: Does the paper discuss both potential positive societal impacts and negative societal impacts of the work performed?
    \item[] Answer: \answerYes{}.
    \item[] Justification: The paper frames the positive impact as improving reproducible audits of reasoning failures and the negative risk as overinterpreting small targeted probes; the limitations and audit contract mitigate this risk by requiring narrow claims and explicit artifact traceability.
    
\item {\bf Safeguards}
    \item[] Question: Does the paper describe safeguards that have been put in place for responsible release of data or models that have a high risk for misuse (e.g., pre-trained language models, image generators, or scraped datasets)?
    \item[] Answer: \answerNA{}.
    \item[] Justification: The paper does not release model weights, scraped datasets, or high-risk generative assets. The release policy in \path{docs/release_packaging_policy.md} excludes raw model responses from the public package by default until model-card response-redistribution terms are reviewed.

\item {\bf Licenses for existing assets}
    \item[] Question: Are the creators or original owners of assets (e.g., code, data, models), used in the paper, properly credited and are the license and terms of use explicitly mentioned and properly respected?
    \item[] Answer: \answerNo{}.
    \item[] Justification: The draft records model identifiers and revisions where available, does not redistribute model weights, and includes a conservative license-status tracker in \path{docs/asset_license_status.md}. Hugging Face metadata now verifies the model license fields used by the experiments, \path{docs/runtime_dependency_license_inventory.md} records a partial isolated-runtime inventory, and \path{docs/official_template_terms_status.md} records unresolved template redistribution terms; the answer remains \answerNo{} until model-card terms relevant to preserved responses, template redistribution terms, unresolved runtime package license entries for any redistributed package contents, and the release license for new project assets are reviewed and recorded.

\item {\bf New assets}
    \item[] Question: Are new assets introduced in the paper well documented and is the documentation provided alongside the assets?
    \item[] Answer: \answerYes{}.
    \item[] Justification: The new task banks, prompt grammar, run configs, analysis summaries, audit rows, artifact traceability rules, and release package boundaries are documented in the appendix and supplemental project files, including task-bank cards, the artifact traceability runbook, the release policy, and the generated public supplement manifest.

\item {\bf Crowdsourcing and research with human subjects}
    \item[] Question: For crowdsourcing experiments and research with human subjects, does the paper include the full text of instructions given to participants and screenshots, if applicable, as well as details about compensation (if any)? 
    \item[] Answer: \answerNA{}.
    \item[] Justification: The paper does not involve crowdsourcing or research with human subjects. The audit labels are part of internal artifact review rather than a human-subjects study.

\item {\bf Institutional review board (IRB) approvals or equivalent for research with human subjects}
    \item[] Question: Does the paper describe potential risks incurred by study participants, whether such risks were disclosed to the subjects, and whether Institutional Review Board (IRB) approvals (or an equivalent approval/review based on the requirements of your country or institution) were obtained?
    \item[] Answer: \answerNA{}.
    \item[] Justification: The work does not involve study participants, personal data, crowdsourcing, or human-subjects experimentation.

\item {\bf Declaration of LLM usage}
    \item[] Question: Does the paper describe the usage of LLMs if it is an important, original, or non-standard component of the core methods in this research? Note that if the LLM is used only for writing, editing, or formatting purposes and does \emph{not} impact the core methodology, scientific rigor, or originality of the research, declaration is not required.
    \item[] Answer: \answerYes{}.
    \item[] Justification: The paper's experiments explicitly evaluate LLM responses from Qwen2.5-7B, Phi-3 Mini, and Mistral-7B, preserve raw responses in the local evidence archive, and separate automatic extraction from audited model-error claims. The default public supplement excludes raw response text and provides sanitized audit rows plus response hashes.

\end{enumerate}

\end{document}